\documentclass[lettersize,journal,onecolumn]{IEEEtran}
\usepackage{amsmath,amsfonts}
\usepackage{array}
\usepackage{booktabs}
\usepackage{tabularx}
\usepackage{graphicx}
\usepackage{xcolor}
\usepackage{enumitem}
\usepackage{placeins}
\usepackage{url}
\usepackage[nocompress]{cite}

\graphicspath{{images/}}
\newcolumntype{P}[1]{>{\raggedright\arraybackslash\hspace{0pt}}p{#1}}

\newcommand{\safeincludegraphics}[2][]{%
\IfFileExists{images/#2}{\includegraphics[#1]{#2}}{\fbox{\parbox{0.9\linewidth}{\centering Missing figure: \texttt{\detokenize{#2}}}}}%
}
\begin{document}

\title{Computational Measurement of Team-Process Phase Dynamics in Collaborative Virtual Reality}

\author{
Qing~Huang,
Jianing~Zhang,
and~Pooja~Pol%
\thanks{Qing Huang, Jianing Zhang, and Pooja Pol are with the
School of Business, Technical University of Applied Sciences Augsburg,
86161 Augsburg, Germany, and also with the
Data Science und Autonome Systeme Technologietransferzentrum (TTZ),
86899 Landsberg am Lech, Germany
(e-mail: qing.huang@tha.de; jianing.zhang@tha.de; pooja.pol@tha.de).
Corresponding author: Qing Huang.}%
}

\maketitle

\begin{abstract}
Collaborative virtual reality (VR) environments make team communication observable as it unfolds, but conventional transcript analyses often summarize entire trials or divide them into fixed temporal windows. Such approaches can obscure changes in team communication and coordination over time. This article presents a computational framework for detecting and interpreting dynamic team-process phases from timestamped dialogue in a collaborative VR game. The framework uses late chunking to generate context-aware transcript representations, aggregates them into temporal chunks, and applies penalized Gaussian-kernel change-point detection to identify semantic transitions in team communication. After boundary detection, term frequency--inverse document frequency (TF--IDF), non-negative matrix factorization (NMF), and representative transcript segments provide structured evidence for phase interpretation. A locally deployed large language model (LLM) uses in-context learning to generate initial interpretations that are subsequently reviewed by humans. Independently recorded interaction logs are then aligned with the detected phases to examine corresponding task-action patterns. The evaluation compares representations, pooling strategies, segmentation methods, parameter settings, reviewed phase interpretations, and phase-aligned interaction profiles. The results show that the framework identifies coherent and interpretable phase structures while preserving traceability to the underlying transcript evidence. The correspondence between transcript-derived phases and interaction behavior further supports their relevance for analyzing collaborative activity. The framework therefore offers a transparent and transferable approach for studying temporal changes in teamwork from timestamped transcripts across collaborative task settings.
\end{abstract}
\begin{IEEEkeywords}
virtual reality collaboration, team communication, team process, change point detection, late chunking, large language models, in-context learning
\end{IEEEkeywords}

\section{Introduction}
\label{sec:introduction}

\IEEEPARstart{C}{ollaborative} virtual reality (VR) training and teamwork platforms provide shared task environments in which communication, coordination, and interaction can be examined as they unfold~\cite{Sparks2025TeamDynamics,Wolfartsberger2020SupportingTeamwork,Lehmann2025TeamworkVR}. Because speech, task actions, and system events can be recorded as temporally aligned traces, VR also offers a useful setting for computational teamwork research~\cite{Goldberg2024AdaptiveCoaching,LehmannWillenbrock2024MultimodalSSP}. Team members must establish shared understanding, coordinate their actions, monitor the task situation, and resolve communication problems during collaboration~\cite{Marks2001TeamProcesses,Cooke2013,Lane2024DigitalWorkplace}. The key analytical challenge is therefore to identify how these communicative and coordination processes change over time rather than merely summarizing the overall amount of interaction~\cite{Sparks2025TeamDynamics,Klonek2025AboutTime}.

Team-process research describes teamwork as a sequence of changing activities, including task interpretation, shared-reference formation, action coordination, and problem response~\cite{Marks2001TeamProcesses,Cooke2013}. Recent temporal perspectives similarly emphasize that team processes and emergent states should be examined as evolving trajectories rather than static aggregates~\cite{Klonek2025AboutTime,Georganta2021ProcessSequences,Li2025ChangeTrajectories,Fyhn2023EmergentStates}. However, raw transcripts do not directly reveal where one collaborative phase ends and another begins. Trial-level summaries remove temporal structure, while fixed temporal windows may split coherent episodes or combine intervals with different communicative functions. This motivates methods that detect variable-length semantic phases from changes in the communication trajectory~\cite{Harrison2022}.

A further challenge is that short VR utterances often depend on surrounding discourse. Expressions such as ``there,'' ``the green one,'' or ``wait'' cannot be interpreted reliably without preceding dialogue and task context. Representing each segment independently may therefore remove information needed to distinguish its local function, which motivates context-preserving approaches such as late chunking~\cite{Gunther2024LateChunking,Merola2025Chunking}. Phase interpretation must also remain traceable to the underlying evidence. Rather than asking an LLM to summarize an entire trial, structured phase-specific evidence can be used to generate candidate interpretations that are subsequently reviewed by humans~\cite{Dong2024ICLSurvey,TopicTag2024,Khandelwal2025,Spain2025LLMTeamTraining}.

\begin{figure}[!t]
    \centering
    \safeincludegraphics[width=0.99\textwidth]{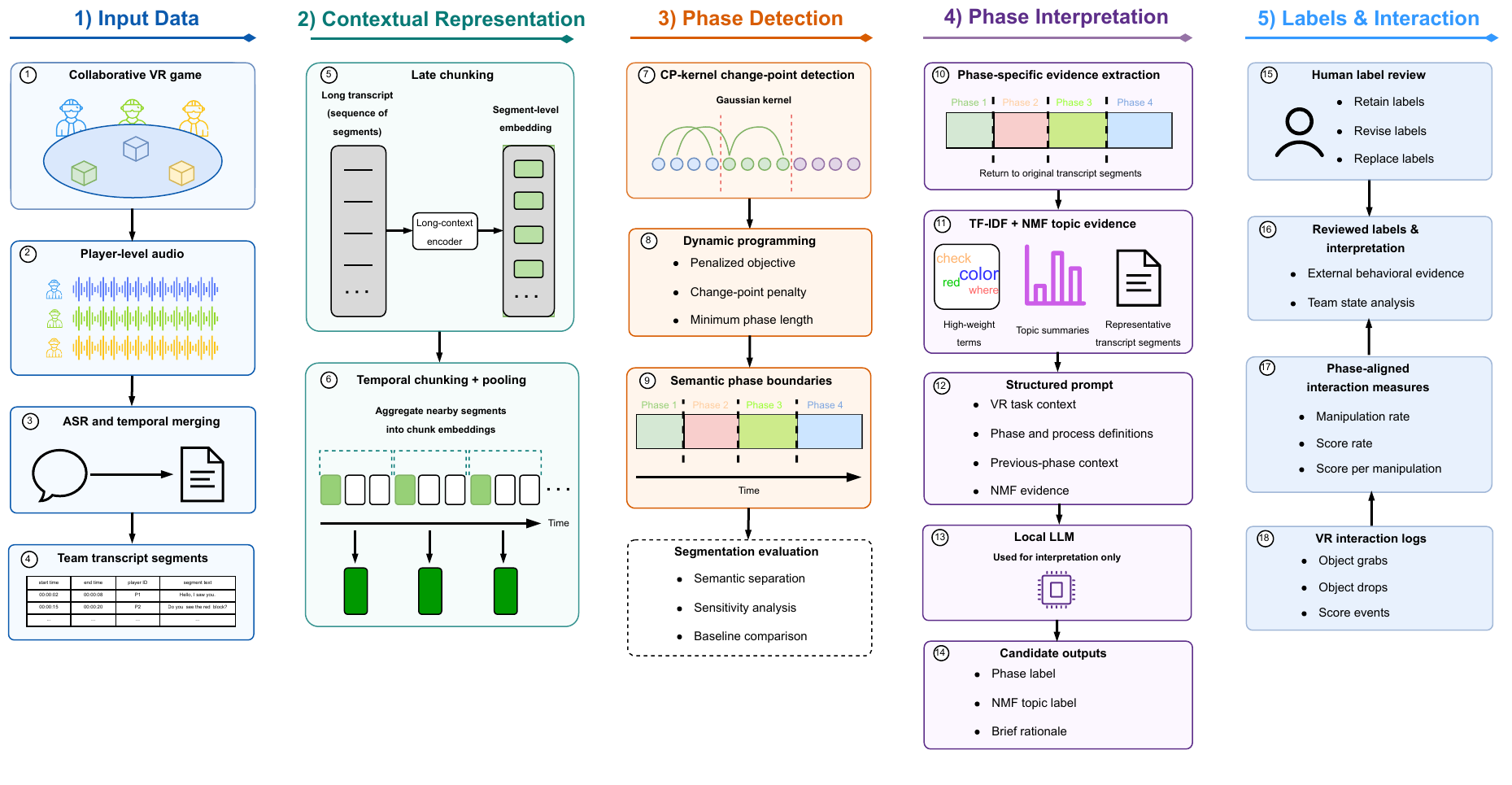}
    \caption{Workflow for semantic phase detection, evidence extraction, and phase interpretation.}
    \label{fig:pipeline_overview}
\end{figure}

This article presents a computational framework for detecting and interpreting semantic phases from timestamped team transcripts. As shown in Fig.~\ref{fig:pipeline_overview}, the framework combines late-chunked contextual embeddings, temporal chunk pooling, kernel-based change-point (CP) detection, structured evidence extraction, LLM-assisted interpretation with in-context learning, human review, and phase-aligned interaction analysis~\cite{Gunther2024LateChunking,JiaDiaz2026,Goldberg2024AdaptiveCoaching}. Semantic boundaries are determined before interpretation so that the labeling process cannot influence their placement. The evaluation examines contextual representation, pooling strategy, segmentation method, and parameter sensitivity, and then assesses the resulting phases through reviewed phase-process labels, non-negative matrix factorization (NMF) topic evidence, and aligned task-action indicators. The framework provides a traceable approach for studying how team communication and coordination develop during collaborative activity.

\section{Related Work and Research Motivation}
\label{sec:related_work}

Research on teamwork emphasizes that team behavior should be analyzed as an evolving process rather than as a static aggregate score~\cite{Marks2001TeamProcesses,Cooke2013}. This view is increasingly important in digital, multimodal, and virtual settings, where emergent states, coordination patterns, and resilience can change over multiple timescales~\cite{Klonek2025AboutTime,Georganta2021ProcessSequences,Li2025ChangeTrajectories,Fyhn2023EmergentStates,Grimm2023TeamResilience,Peifer2021TeamFlowVirtualTeams,Lane2024DigitalWorkplace,LehmannWillenbrock2024MultimodalSSP,Goldberg2024AdaptiveCoaching}. For human--machine systems, training feedback and after-action review require interpretable information about when and how these processes change.

Existing computational analyses often operate either at the transcript-segment level or at the trial level. Segment-level labels can be too local to describe sustained team-process episodes, whereas trial-level summaries collapse meaningful transitions. The needed intermediate unit is a semantically coherent phase that is longer than a single transcript segment, shorter than a full task episode, and variable in duration according to the team's communication dynamics.

Collaborative VR dialogue differs from well-formed documents because it contains short turns, deictic expressions, simultaneous player perspectives, task-specific object references, and local repair sequences. Late chunking is relevant because it encodes a longer context first and then derives local segment representations, allowing short segments to inherit discourse information~\cite{Gunther2024LateChunking,Merola2025Chunking}. For boundary estimation, kernel change-point detection is suitable because it evaluates changes in the distribution of a semantic trajectory rather than relying only on adjacent local similarity~\cite{Hearst1997TextTiling,Harrison2022,JiaDiaz2026}.

Interpretation remains separate from boundary detection. A boundary set does not specify what communicative function a phase serves. Recent work on in-context learning and LLM-assisted communication analysis suggests a useful division of labor in which computational methods extract structured evidence, LLMs convert that evidence into reviewable candidate labels, and human reviewers assess whether the labels are supported by the evidence~\cite{Dong2024ICLSurvey,Chang2024SpeechICL,TopicTag2024,Khandelwal2025,Spain2025LLMTeamTraining,Raut2025MicroBehaviors}. The present study follows this evidence-first logic and treats LLM use as LLM-assisted evidence-grounded annotation within a human-reviewed workflow.

\section{Research Questions and Claim Boundaries}
\label{sec:research_questions}

The evaluation follows four questions aligned with the measurement pipeline. RQ1 examines whether late chunking improves semantic separation relative to standard segment embeddings. RQ2 examines whether CP-kernel segmentation produces higher semantic separation than fixed-width and random-boundary reference partitions. RQ3 examines sensitivity to the embedding model, chunk pooling, kernel bandwidth, and change-point penalty. RQ4 examines whether reviewed phase-process labels can be assigned to the detected phases, whether these labels are supported by interpreted NMF topic evidence, and whether the phases can be aligned with VR interaction logs for cross-session comparisons of communication patterns and task-action profiles.

\textbf{Claim boundary.} The paper evaluates transcript-semantic structure, reviewed interpretability, and alignment with task-action and performance measures. The detected phases are analytic units for examining team communication and coordination in collaborative VR rather than direct measurements of latent psychological states. External validation through trainer ratings, participant self-reports, and independent behavioral coding remains future work.

\section{Data and Preprocessing}
\label{sec:data}

\subsection{VR Game Context and Transcript Data}
\label{subsec:study_context}

The study was conducted in a collaborative VR training game designed around communication and coordination rather than individual dexterity. Participants entered a shared virtual environment as visually similar avatars and had to move task-relevant resources to target locations under time pressure. The task required team members to identify player-specific attributes, establish mutual visibility, localize resources, coordinate object manipulation, and repair misunderstandings during the ongoing activity. Each team completed two VR-game trials. Between trials, the group left VR for a facilitated reflection and strategy discussion, then returned for a second attempt. This paired design makes it possible to compare an initial encounter with the task and a later attempt after explicit team reflection. All participants provided written informed consent before participation, including consent for the recording and analysis of audio, transcript, and VR interaction data.

The data consist of timestamped German speech transcripts and timestamped VR interaction logs from these trials. Color terms in this article refer only to VR-game properties, such as object colors or color-coded player roles, and should not be interpreted as demographic, personal, or social descriptors. The dataset contains 15 teams, each of which completed two trials, resulting in 30 team-session transcripts. Player-level audio was transcribed with a Whisper-based automatic speech recognition (ASR) pipeline~\cite{Radford2023Whisper}. A \emph{transcript segment} is the minimal textual unit used in the analysis and consists of one timestamped ASR output segment from a player-level audio recording. Because each participant had a separate recording channel, the player identifier was inherited from the source channel rather than inferred through diarization. The $i$th segment is
\begin{equation}
    \sigma_i=(s_i,e_i,c_i,x_i),
\end{equation}
where $s_i$ and $e_i$ denote the start and end timestamps, $c_i$ denotes the player identifier, and $x_i$ denotes the transcript text. A team trial is represented as the ordered sequence
\begin{equation}
    S=\{\sigma_1,\sigma_2,\ldots,\sigma_N\},
\end{equation}
where $N$ is the number of usable timestamped ASR segments.

Empty, non-speech, and unusable ASR fragments were removed, and silent intervals were not converted into pseudo-segments. Transcript text and timestamp alignment were manually checked. Because the focus of this study is semantic phase detection from timestamped transcripts rather than speech-to-text performance, ASR quality is not evaluated as a separate research objective, and the effect of transcription errors on the downstream results is not examined here.

\subsection{Contextual Segment Embeddings and Temporal Chunks}
\label{subsec:late_chunking_method}

Three analysis units are used in the framework. A \emph{transcript segment} is the timestamped automatic speech recognition (ASR) unit defined above. A \emph{contextual segment embedding} represents the semantic content of that segment together with its surrounding discourse. A \emph{temporal chunk} aggregates nearby contextual segment embeddings for change-point detection. Transcript segments retain the original textual evidence, while temporal chunks provide a more stable representation for boundary estimation.

For late chunking, neighboring transcript segments are concatenated in temporal order into a \emph{transcript block} up to the encoder context limit. Let
\begin{equation}
    T^{(b)}=(\tau_1,\tau_2,\ldots,\tau_L)
\end{equation}
denote the token sequence of transcript block $b$, where $\tau_\ell$ is the $\ell$th token and $L$ is the number of tokens in the block. A long-context encoder $f_\theta$ maps this sequence to contextual token representations.
\begin{equation}
    (\boldsymbol{\vartheta}_1,\boldsymbol{\vartheta}_2,\ldots,\boldsymbol{\vartheta}_L)
    = f_\theta(T^{(b)}).
\end{equation}
Here, $\boldsymbol{\vartheta}_\ell$ denotes the contextual representation of token $\tau_\ell$, and $\theta$ denotes the parameters of the embedding model. The embedding of transcript segment $\sigma_i$ is obtained by pooling the contextual token representations aligned with that segment.
\begin{equation}
\begin{aligned}
    \mathbf{x}_i
    &=
    \operatorname{MeanPool}_{\mathrm{token}}\bigl(\{\boldsymbol{\vartheta}_{\ell} \mid \tau_{\ell}\in\sigma_i\}\bigr),\\
    \tilde{\mathbf{x}}_i
    &=
    \frac{\mathbf{x}_i}{\|\mathbf{x}_i\|_2}.
\end{aligned}
\end{equation}
In these expressions, $\mathbf{x}_i$ is the pooled embedding of transcript segment $\sigma_i$, and $\tilde{\mathbf{x}}_i$ is its $L_2$-normalized form. Token representations are combined using mean pooling. Normalization places all segment embeddings on a comparable scale for cosine-based evaluation and kernel-based distance computation.

When a transcript exceeds the encoder context limit, it is divided into overlapping transcript blocks. If a segment appears in more than one block, its resulting embeddings are combined before normalization. This procedure preserves broader conversational context while maintaining alignment with the original ASR timestamps. The standard embedding condition processes each transcript segment independently and serves as the comparison condition for evaluating the effect of late chunking.

Change-point detection is not applied directly to individual transcript segments. These segments are often short, unevenly distributed over time, and semantically ambiguous without local context. Direct segmentation at this level would therefore be sensitive to local transcription and timing fluctuations. Instead, nearby contextual segment embeddings are aggregated into temporal chunks of width $\Delta$.
\begin{equation}
\begin{aligned}
    I_k
    &=
    [t_0+(k-1)\Delta,\;t_0+k\Delta),\\
    C_k
    &=
    \{\sigma_i:s_i\in I_k\},\\
    \mathbf{u}_k
    &=
    \operatorname{Pool}_{\mathrm{chunk}}\bigl(\{\tilde{\mathbf{x}}_i \mid \sigma_i\in C_k\}\bigr),\\
    \mathbf{z}_k
    &=
    \frac{\mathbf{u}_k}{\|\mathbf{u}_k\|_2},
    \qquad \|\mathbf{u}_k\|_2>0.
\end{aligned}
\label{eq:temporal_chunk}
\end{equation}
Here, $I_k$ denotes the $k$th temporal interval, $t_0$ is the trial start time, and $\Delta$ is the temporal chunk width. The set $C_k$ contains all transcript segments whose start timestamps fall within $I_k$. The vector $\mathbf{u}_k$ is the pooled chunk representation and $\mathbf{z}_k$ is its $L_2$-normalized form. Thus, after mean, maximum, minimum, or mean--maximum pooling, each nonzero chunk embedding is divided by its Euclidean norm so that its length equals one. This normalization is applied consistently across all chunk-pooling strategies before kernel computation. The ordered sequence
\begin{equation}
    \mathbf{Z}=\{\mathbf{z}_1,\mathbf{z}_2,\ldots,\mathbf{z}_M\}
\end{equation}
forms the semantic trajectory used for boundary estimation, where $M$ is the number of non-empty temporal chunks. Chunks without usable transcript segments are excluded rather than imputed. After the boundaries are detected, phase interpretation and human review return to the original transcript segments. Temporal chunking therefore stabilizes boundary estimation while preserving access to the underlying textual evidence.

\section{Method}
\label{sec:method}

For each trial, the computation follows the steps below.
\begin{enumerate}
    \item Construct the ordered transcript sequence $S=\{\sigma_1,\ldots,\sigma_N\}$ from player-level ASR segments.
    \item Compute one normalized embedding $\tilde{\mathbf{x}}_i$ for each transcript segment using either standard segment embedding or late-chunked contextual embedding.
    \item Aggregate the segment embeddings into temporal chunk vectors $\mathbf{Z}=\{\mathbf{z}_1,\ldots,\mathbf{z}_M\}$ using the selected pooling rule.
    \item Compute the Gaussian kernel matrix over $\mathbf{Z}$ and the interval costs $\widehat{C}(s,e)$ for admissible chunk intervals.
    \item Estimate the CP-kernel boundaries $\widehat{\boldsymbol{\tau}}$ by minimizing the penalized objective with exact dynamic programming.
    \item Map the detected phase intervals back to the original transcript segments.
    \item Convert each fixed phase interval into an evidence packet for LLM-assisted interpretation, human review, and phase-aligned interaction analysis.
\end{enumerate}

\subsection{Gaussian-Kernel Change-Point Objective}
\label{subsec:kcpd}

Each trial is treated as an ordered trajectory of chunk vectors $\mathbf{z}_k$. The objective is to identify internally coherent intervals separated by changes in the embedding distribution. A Gaussian radial basis function (RBF) kernel is therefore used to compare chunks.
\begin{equation}
    k(\mathbf{z}_i,\mathbf{z}_j)
    =
    \exp\left(-\gamma\|\mathbf{z}_i-\mathbf{z}_j\|_2^2\right).
\label{eq:rbf_kernel}
\end{equation}
Here $\gamma$ controls the sensitivity of the Gaussian kernel to distances between chunk embeddings and $\|\cdot\|_2$ is the Euclidean norm. The Gaussian kernel captures changes in the geometry of the embedding distribution, including shifts in location, dispersion, and local density. Such changes can reflect transitions from exploratory discussion to focused coordination even when task vocabulary remains similar.

For a candidate interval $[s,e]$, the empirical kernel cost is
\begin{equation}
\begin{aligned}
    \widehat{C}(s,e)
    &=
    \sum_{t=s}^{e} k(\mathbf{z}_t,\mathbf{z}_t)\\
    &\quad
    -\frac{1}{e-s+1}
    \sum_{i=s}^{e}\sum_{j=s}^{e}
    k(\mathbf{z}_i,\mathbf{z}_j).
\end{aligned}
\label{eq:kernel_cost}
\end{equation}
Here $\widehat{C}(s,e)$ is the cost of representing chunks $s$ through $e$ as one phase, and $e-s+1$ is the number of chunks in the interval. A coherent interval has a lower cost because the within-interval kernel similarities are high. An interval that spans a semantic shift is more heterogeneous and therefore receives a higher cost.

The estimated change points minimize the penalized objective
\begin{equation}
\begin{aligned}
    \widehat{\boldsymbol{\tau}}
    &=
    \arg\min_{\boldsymbol{\tau}} J(\boldsymbol{\tau}),\\
    J(\boldsymbol{\tau})
    &=
    \sum_{r=1}^{K+1}
    \widehat{C}(\tau_{r-1}+1,\tau_r)+\beta K,\\
    \boldsymbol{\tau}
    &:\;0=\tau_0<\cdots<\tau_K<\tau_{K+1}=M.
\end{aligned}
\label{eq:cp_objective}
\end{equation}
In this objective, $\widehat{\boldsymbol{\tau}}$ is the estimated boundary sequence, $J(\boldsymbol{\tau})$ is the total segmentation cost, $K$ is the number of detected change points, $K+1$ is the number of resulting phases, $\tau_r$ is the end chunk of phase $r$, $M$ is the number of temporal chunks in the trial, and $\beta$ penalizes additional boundaries. A minimum phase length in chunks prevents uninterpretable micro-phases. The default kernel scale is derived from the median pairwise squared distance,
\begin{equation}
    \gamma_0 = \frac{1}{d_{\mathrm{med}}},
\label{eq:gamma_base}
\end{equation}
where $d_{\mathrm{med}}$ is the median of the nonzero pairwise squared distances between chunk embeddings. Sensitivity analysis uses
\begin{equation}
    \gamma=\alpha\gamma_0,
\label{eq:gamma_multiplier}
\end{equation}
where $\alpha$ is the gamma multiplier reported in the sensitivity plots.

Sensitivity analysis uses late-chunked \texttt{jina-v2-de}, mean pooling, $\alpha=1$, $\beta=1$, and $m_{\min}=1$ as the reference configuration. Embedding model, embedding mode, pooling rule, gamma multiplier, and penalty are varied separately. Both $\alpha$ and $\beta$ are evaluated over $\{0.01,0.05,0.1,0.5,1,5,10\}$. For each condition, boundaries are recomputed and evaluated with the same semantic-separation score. LLM labels and interaction logs are not used for parameter selection.

In the reported experiments, the penalized objective in (\ref{eq:cp_objective}) is solved by exact dynamic programming over the ordered chunk sequence. Let $F(t)$ denote the minimum segmentation cost for the prefix ending at chunk $t$. With minimum phase length $m_{\min}$ and admissible predecessor set $\mathcal{A}(t)=\{s:0\le s<t,\ t-s\ge m_{\min}\}$, the Bellman recursion is
\begin{equation}
\begin{aligned}
F(t) &=
\min_{s\in\mathcal{A}(t)}
\Bigl[
F(s)+\widehat{C}(s+1,t)\\
&\hspace{2.6em}
+\beta\,\mathbb{I}(s>0)
\Bigr],\\
F(0) &= 0.
\end{aligned}
\label{eq:exact_dp}
\end{equation}
Here $\mathbb{I}(s>0)$ denotes an indicator that equals one only when the candidate segment does not start at the beginning of the trial. Thus, the first phase is penalty-free, whereas every additional boundary contributes $\beta$ to the objective. For each endpoint $t$, the algorithm evaluates all admissible predecessor indices $s$, stores the best predecessor, and reconstructs the optimal boundary sequence by backtracking from $t=M$. This exact dynamic-programming formulation is feasible because temporal chunking keeps the number of analysis units per trial moderate.

\subsection{Segmentation Methods and Evaluation Metric}
\label{subsec:segmentation_baselines_segment_level}

The empirical comparison considers three segmentation methods applied to the same semantic chunk trajectory. The proposed method is CP-kernel segmentation, which minimizes the penalized kernel objective in (\ref{eq:cp_objective}) and places boundaries where the semantic distribution of the chunk sequence changes. It is therefore the only method that uses the semantic objective directly for boundary placement.

The first reference method is fixed-width segmentation. It uses the same number of phases as the CP-kernel solution, $P=K+1$, but replaces adaptive boundary placement with an equal-width partition.
\begin{equation}
    \tau_r^{\mathrm{fw}} = \left\lfloor \frac{rM}{P} \right\rfloor, \qquad r=1,\dots,P-1,
\label{eq:fixed_width}
\end{equation}
where $\tau_r^{\mathrm{fw}}$ is the $r$th fixed-width boundary, $M$ is the number of temporal chunks, and $P$ is the number of phases. This baseline tests whether a regular temporal grid is sufficient to recover interpretable semantic phases.

The second reference method is random-boundary segmentation. It preserves the same number of change points and the same minimum phase-length constraint as the CP-kernel solution, but it discards the semantic objective used for boundary placement. A random boundary sequence satisfies
\begin{equation}
0=\tau^{\mathrm{rand}}_0
<\tau^{\mathrm{rand}}_1
<\cdots
<\tau^{\mathrm{rand}}_{P-1}
<\tau^{\mathrm{rand}}_P=M,
\end{equation}
and every resulting phase must satisfy
\begin{equation}
\tau^{\mathrm{rand}}_r-\tau^{\mathrm{rand}}_{r-1}
\geq m_{\min},
\qquad r=1,\ldots,P.
\label{eq:admissible_boundary}
\end{equation}
Here, $m_{\min}$ is the minimum allowed phase length in chunks. For each session, 1000 admissible random boundary sequences are sampled under this constraint using a fixed pseudo-random generator with seed 42. Each random segmentation is matched to the CP-kernel output in both the number of change points and the minimum phase-length constraint. The random-reference result is summarized by the mean semantic-separation score across the 1000 repetitions together with an empirical 95\% confidence interval. Using the same seeded random-number generator makes the baseline reproducible for identical inputs. This baseline therefore functions as a chance-level control that preserves segmentation granularity while removing optimized boundary placement.

These three methods are compared with the same semantic-separation score. Because boundary estimation is performed on temporal chunks, semantic separation is evaluated primarily on the chunk embeddings used by the segmentation procedure. Let $C_p$ denote the set of temporal chunks assigned to phase $p$. For phases containing at least two chunks, within-phase similarity is the average pairwise cosine similarity among chunk embeddings in $C_p$.
\begin{equation}
    W_p^{\mathrm{chunk}}
    =
    \frac{2}{|C_p|(|C_p|-1)}
    \sum_{\substack{i,j\in C_p\\i<j}}
    \cos(\mathbf{z}_i,\mathbf{z}_j),
\label{eq:within_similarity}
\end{equation}
where $W_p^{\mathrm{chunk}}$ is the within-phase chunk similarity for phase $p$ and $\cos(\mathbf{z}_i,\mathbf{z}_j)$ is the cosine similarity between the normalized chunk embeddings.

Between-phase similarity compares the chunks in $C_p$ with all chunks outside that phase.
\begin{equation}
    B_p^{\mathrm{chunk}}
    =
    \frac{1}{|C_p|\,|C_{\neg p}|}
    \sum_{i\in C_p}\sum_{j\in C_{\neg p}}
    \cos(\mathbf{z}_i,\mathbf{z}_j),
\label{eq:between_similarity}
\end{equation}
where $C_{\neg p}$ denotes all temporal chunks in the same session that are not assigned to phase $p$. The corresponding chunk-level phase separation score is
\begin{equation}
    D_p^{\mathrm{chunk}}
    =
    W_p^{\mathrm{chunk}}-B_p^{\mathrm{chunk}}.
\label{eq:phase_separation}
\end{equation}
A larger value indicates that chunks within the same phase are more semantically similar to one another than to chunks outside that phase.

If a detected phase contains only one temporal chunk, the pairwise term in $W_p^{\mathrm{chunk}}$ is undefined. In this case, the score is computed from the original transcript-segment embeddings assigned to that phase rather than discarding the session. Let $S_p$ denote the set of transcript segments mapped to phase $p$. The fallback within-phase and between-phase similarities are
\begin{equation}
    W_p^{\mathrm{seg}}
    =
    \frac{2}{|S_p|(|S_p|-1)}
    \sum_{\substack{i,j\in S_p\\i<j}}
    \cos(\tilde{\mathbf{x}}_i,\tilde{\mathbf{x}}_j),
\label{eq:within_similarity_segment}
\end{equation}
\begin{equation}
    B_p^{\mathrm{seg}}
    =
    \frac{1}{|S_p|\,|S_{\neg p}|}
    \sum_{i\in S_p}\sum_{j\in S_{\neg p}}
    \cos(\tilde{\mathbf{x}}_i,\tilde{\mathbf{x}}_j),
\label{eq:between_similarity_segment}
\end{equation}
with
\begin{equation}
    D_p^{\mathrm{seg}}
    =
    W_p^{\mathrm{seg}}-B_p^{\mathrm{seg}}.
\label{eq:phase_separation_segment}
\end{equation}
The final phase-level score is therefore
\begin{equation}
D_p=
\begin{cases}
D_p^{\mathrm{chunk}}, & |C_p|\geq 2,\\
D_p^{\mathrm{seg}}, & |C_p|=1 \text{ and } |S_p|\geq 2.
\end{cases}
\label{eq:phase_separation_combined}
\end{equation}
The session score is the mean of all valid phase-level scores. If a session collapses to a single detected phase, its score is set to zero because no between-phase contrast exists. For the reference methods, the phase count is matched to the CP-kernel solution for the same embedding generation method. If a valid reference segmentation cannot be constructed under this constraint, the session score is also set to zero.

\subsection{Evidence Extraction and Reviewed Interpretation}
\label{subsec:evidence_labeling_review_method}

After boundaries are fixed, the pipeline returns to transcript evidence inside each phase. This step is deliberately separated from boundary estimation. Evidence extraction and labeling do not move, add, or remove change points. TF--IDF terms and NMF topics summarize recurrent lexical patterns, while representative transcript segments preserve direct qualitative evidence. The NMF model provides topic weights and topic-specific lexical evidence. The labels reported later are interpreted NMF topic labels, not raw lists of high-weight topic terms. For phase $p$, the phase-term matrix is approximated as
\begin{equation}
\begin{aligned}
    \mathbf{V}^{(p)}&\approx\mathbf{W}^{(p)}\mathbf{H}^{(p)},\\
    \mathbf{W}^{(p)},\mathbf{H}^{(p)}&\geq 0,
\end{aligned}
\label{eq:nmf}
\end{equation}
where $\mathbf{V}^{(p)}$ is the TF--IDF-weighted matrix, $\mathbf{W}^{(p)}$ contains segment-topic weights, and $\mathbf{H}^{(p)}$ contains topic-term weights. High-weight terms and representative segments form the evidence base for phase interpretation.

A locally deployed \texttt{qwen3.5:9b} model converts each structured evidence packet into an initial phase interpretation. The LLM serves as an annotation assistant rather than an autonomous evaluator, reducing the need to formulate every label from scratch. In-context examples provide the VR task context, expected abstraction level, and previous-phase information, thereby producing more consistent, task-grounded labels than unconstrained summarization would.

The interpretation record contains a predefined phase-process label for cross-session comparison and interpreted NMF topic labels as supporting semantic evidence. The original transcript evidence is in German, while the reported labels are provided in English for readability. The LLM generates an initial phase-process label and interpreted NMF topic labels from the structured evidence packet. Human review then verifies whether these candidate labels are supported by the phase-specific transcript evidence and task context. Reviewers may retain, revise, or replace the proposed labels, but they do not modify the detected boundaries. The reported outputs are therefore human-reviewed labels from an LLM-assisted annotation process that reduces the need to formulate every label from scratch.

Figure~\ref{fig:prompt_template} shows the prompt template used for the initial LLM-assisted phase interpretation. The prompt combines task-level context with phase-specific evidence, including TF--IDF terms, NMF topic evidence, representative utterances, and information about the preceding phase. It explicitly instructs the model to interpret the complete detected phase rather than individual utterances and to leave the previously detected phase boundaries unchanged. The team-process function is selected from the predefined codebook, while the phase-specific focus provides a more concrete description of the communication occurring within the phase. Requiring a short evidence-based justification and multiple candidate labels makes the initial interpretation easier to inspect during subsequent human review.

\begin{figure}[!htbp]
\centering
\setlength{\fboxsep}{8pt}
\fcolorbox{black!45}{black!3}{%
\begin{minipage}{0.94\linewidth}
\footnotesize
\textbf{Prompt template for context-grounded phase labeling}

\medskip
\textbf{Role and task.} You are an expert in German-language team communication analysis, VR collaboration, and qualitative topic labeling. You will receive evidence from one detected communication phase in a VR-based team collaboration task. Phase boundaries were detected before this labeling step. Your task is to generate concise, academically usable English labels that describe the dominant communicative function of the complete phase. Do not label individual utterances and do not create or modify phase boundaries.

\medskip
\textbf{Context inputs.}
\begin{itemize}
  \item \textbf{Scenario context:} \{scenario\_context\}
  \item \textbf{Team-process codebook and inter-label clarifications:} \{team\_process\_codebook\}
  \item \textbf{Previous phase context:} \{previous\_phase\_summary\}
\end{itemize}

\textbf{Current phase evidence.}
\begin{itemize}
  \item \textbf{Phase ID and time interval:} \{phase\_id\}; \{phase\_start\}--\{phase\_end\}
  \item \textbf{Top TF--IDF terms and n-grams:} \{phase\_top\_terms\}
  \item \textbf{NMF topic evidence:} \{nmf\_topics\}
  \item \textbf{Representative utterances:} \{representative\_utterances\}
\end{itemize}

\textbf{Labeling criteria.}
\begin{enumerate}[leftmargin=1.5em,itemsep=0.2em,topsep=0.2em]
  \item \textbf{Evidence grounding:} base the interpretation on the representative utterances and topic evidence; use NMF as within-phase evidence, not as separate phase labels.
  \item \textbf{Coverage:} label the dominant communicative function of the complete phase, rather than one salient utterance, object, colour, player, or location.
  \item \textbf{Discrimination:} distinguish the phase from adjacent phases; if the same label is retained, state the concrete difference in focus.
  \item \textbf{Specificity and style:} avoid generic labels such as ``communication'', ``teamwork'', ``discussion'', or ``interaction''. Use natural English suitable for a scientific report; prefer labels of 2--5 words.
\end{enumerate}

\textbf{Required output format.}
\begin{quote}
\ttfamily\footnotesize
Team-process function: One codebook category that best captures the dominant function.\\
Phase-specific focus: One short English phrase describing the concrete focus of this phase.\\
Reasoning: One short sentence linking the label to transcript evidence.\\
Candidate labels: ...\\
Final label: ...
\end{quote}
\end{minipage}}
\caption{Context-grounded prompt template for LLM-assisted phase labeling using task context and phase-specific transcript evidence.}
\label{fig:prompt_template}
\end{figure}

\section{Results}
\label{sec:results}

The results examine representation choice, pooling strategy, parameter sensitivity, segmentation baselines, detected phase structure, and evidence-grounded interpretation. All reported analyses use 30-second temporal chunks. In each sensitivity analysis, only the parameter under examination is varied, while all other parameters remain fixed at their final selected values.

\subsection{Representation Choice and the Effect of Late Context}
\label{subsec:embedding_model_selection}

\begin{figure}[!htbp]
    \centering
    \safeincludegraphics[width=0.5\columnwidth]{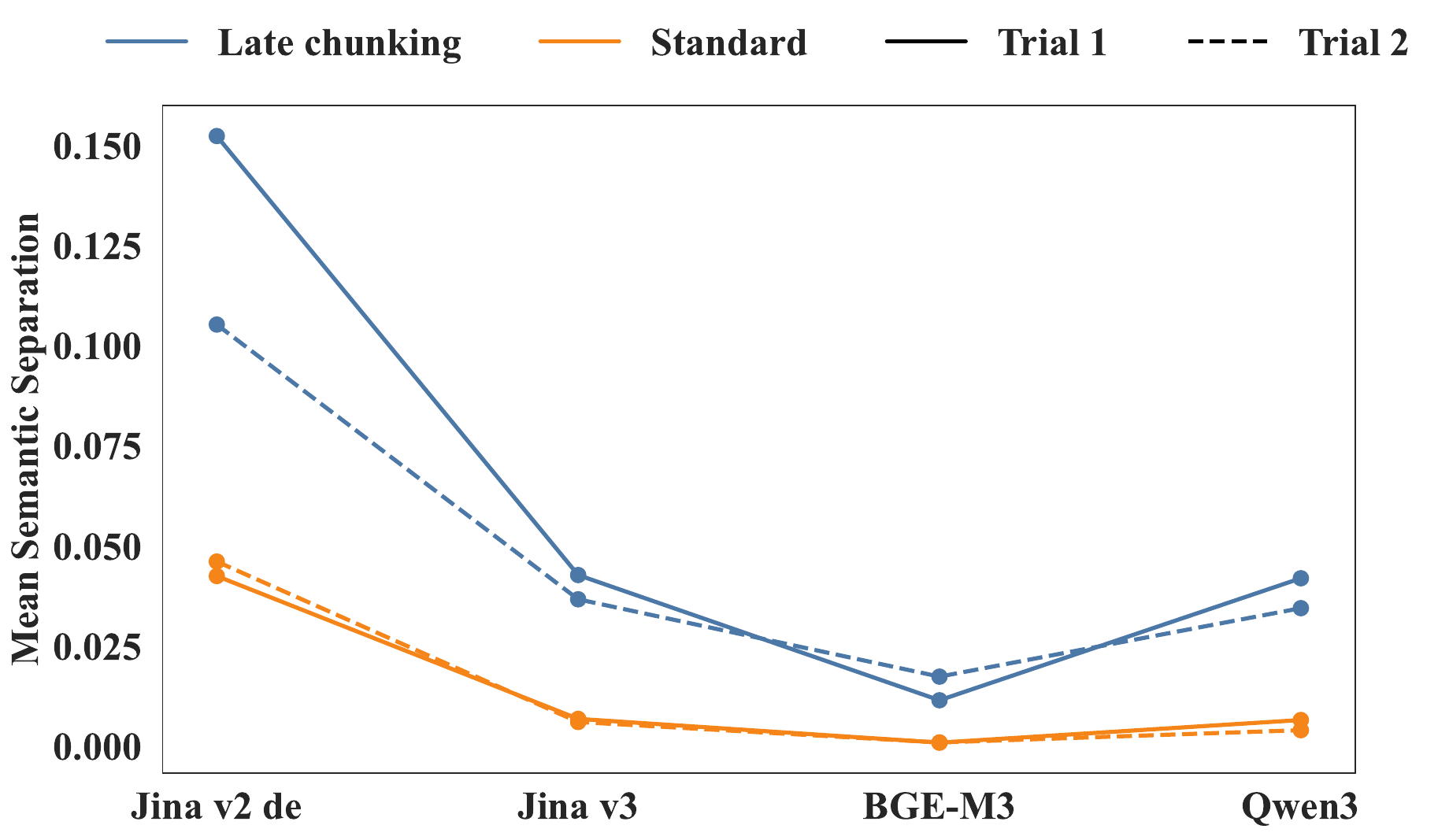}
    \caption{Semantic separation across embedding models and embedding modes using mean pooling.}
    \label{fig:embedding_model_separation}
\end{figure}

Figure~\ref{fig:embedding_model_separation} compares late-chunked and standard embeddings across the tested models using mean pooling and the same semantic-separation metric. Across all tested models, late chunking produces higher semantic separation than standard embedding, indicating that contextualization has a stronger influence on the representation than model choice alone. Within the late-chunked condition, \texttt{jina-v2-de} achieves the highest semantic separation in both trials. These results show that preserving conversational context is the primary factor, while selecting a language-appropriate embedding model provides an additional benefit for downstream phase detection.

\subsection{Pooling Strategy and Chunk-Trajectory Stability}
\label{subsec:pooling_strategy_results}

\begin{figure}[!htbp]
    \centering
    \safeincludegraphics[width=0.5\columnwidth]{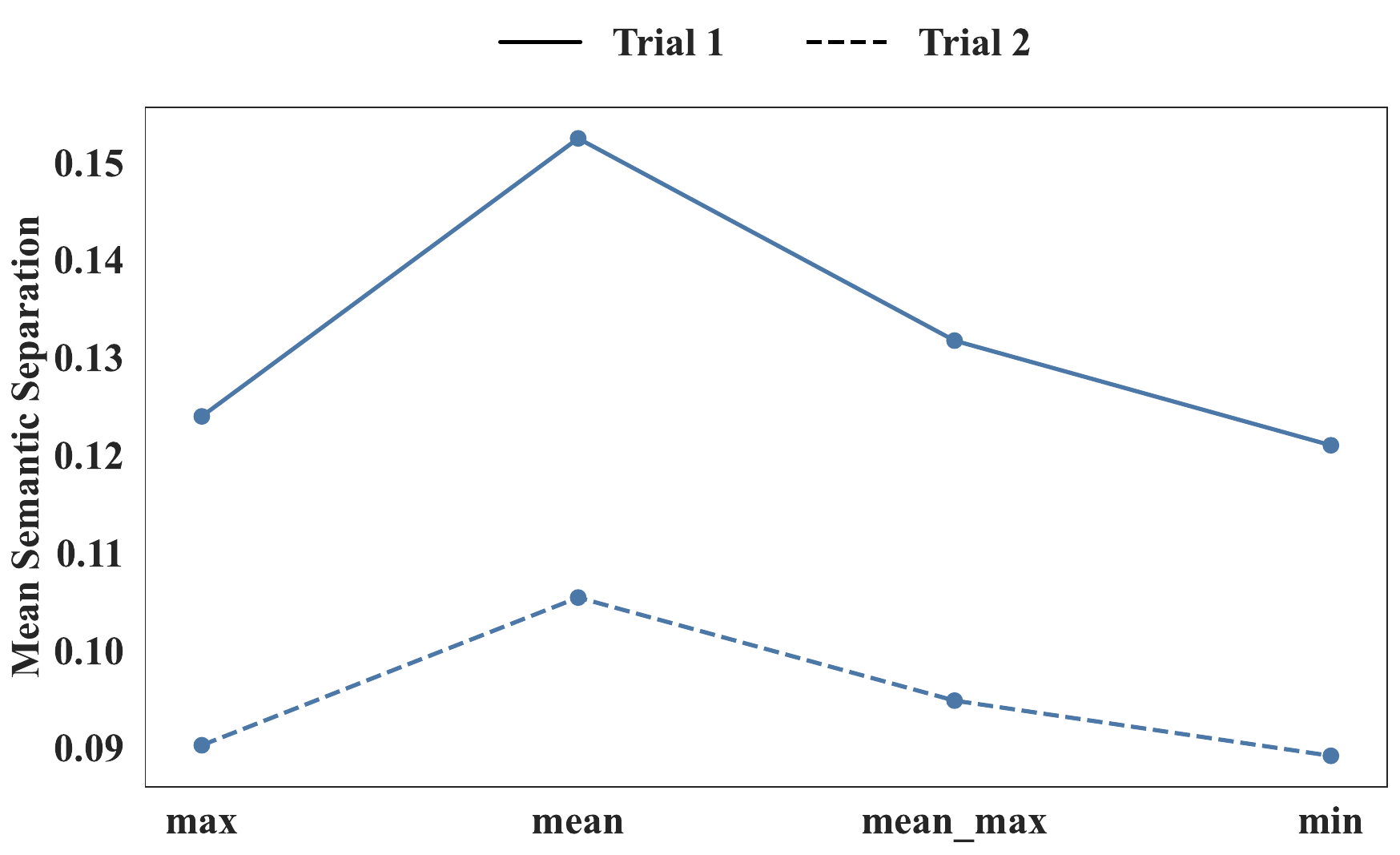}
    \caption{Semantic separation across temporal chunk-pooling strategies using late-chunked \texttt{jina-v2-de} embeddings.}
    \label{fig:pooling_strategy_comparison}
\end{figure}

Figure~\ref{fig:pooling_strategy_comparison} compares chunk-pooling strategies under the late-chunked \texttt{jina-v2-de} representation using the same downstream change-point kernel configuration. Mean pooling produces the highest semantic separation in both trials. A temporal chunk can contain several complementary transcript segments that jointly express the local communication function. Mean pooling captures the central semantic pattern of this exchange, whereas max and min pooling can overemphasize individual segments and mean--max pooling can retain part of this sensitivity. Because temporal pooling is used only for boundary estimation and interpretation returns to the original transcript segments, mean pooling stabilizes the semantic trajectory without allowing a single segment to dominate it.

\subsection{Sensitivity to the Gaussian-Kernel Scale and Change-Point Penalty}
\label{subsec:gamma_sensitivity}

\begin{figure}[!htbp]
    \centering
    \safeincludegraphics[width=0.5\columnwidth]{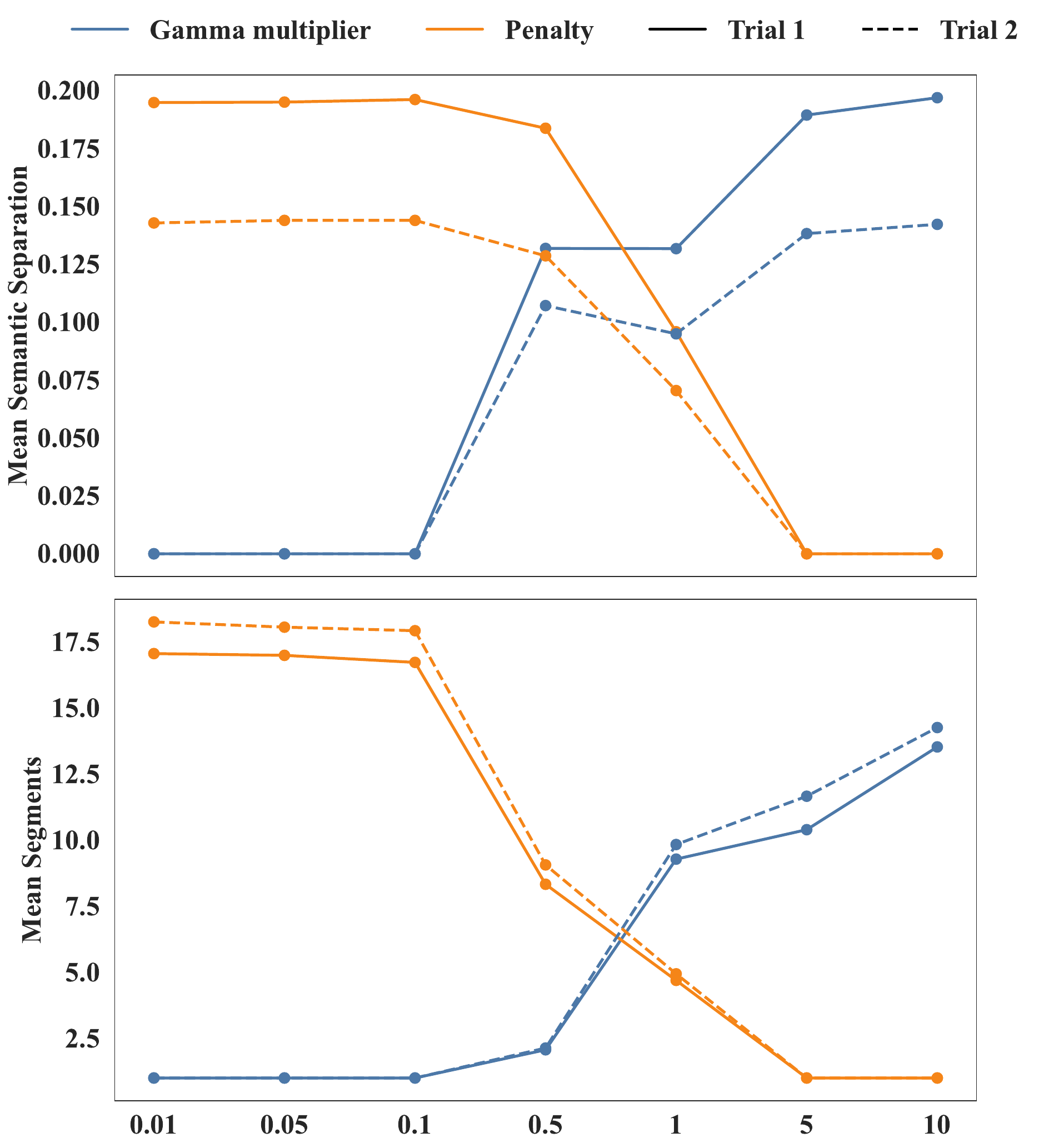}
    \caption{Semantic separation and mean number of detected phases across gamma multipliers and change-point penalties.}
    \label{fig:gamma_penalty_sensitivity}
\end{figure}

Figure~\ref{fig:gamma_penalty_sensitivity} presents the sensitivity of semantic separation and mean phase count to the gamma multiplier and change-point penalty under late-chunked \texttt{jina-v2-de} with mean pooling. Very small gamma multipliers make the Gaussian kernel less sensitive to embedding differences, causing chunks to appear more similar and reducing the number of detected transitions. Larger gamma multipliers make local semantic differences more visible and increase semantic separation, but overly large values can fragment transcripts into many short phases. The penalty controls the opposing tendency. Small penalties permit more boundaries and can increase apparent semantic separation, whereas large penalties suppress boundaries and can collapse the phase structure. The final setting of $\alpha=1$ and $\beta=1$ provides clear semantic separation without excessive fragmentation or phase collapse. Both trials exhibit the same overall trade-off.

\subsection{Baseline Comparison of Adaptive and Reference Boundaries}
\label{subsec:baseline_comparison_results}

\begin{figure}[!htbp]
    \centering
    \safeincludegraphics[width=0.5\columnwidth]{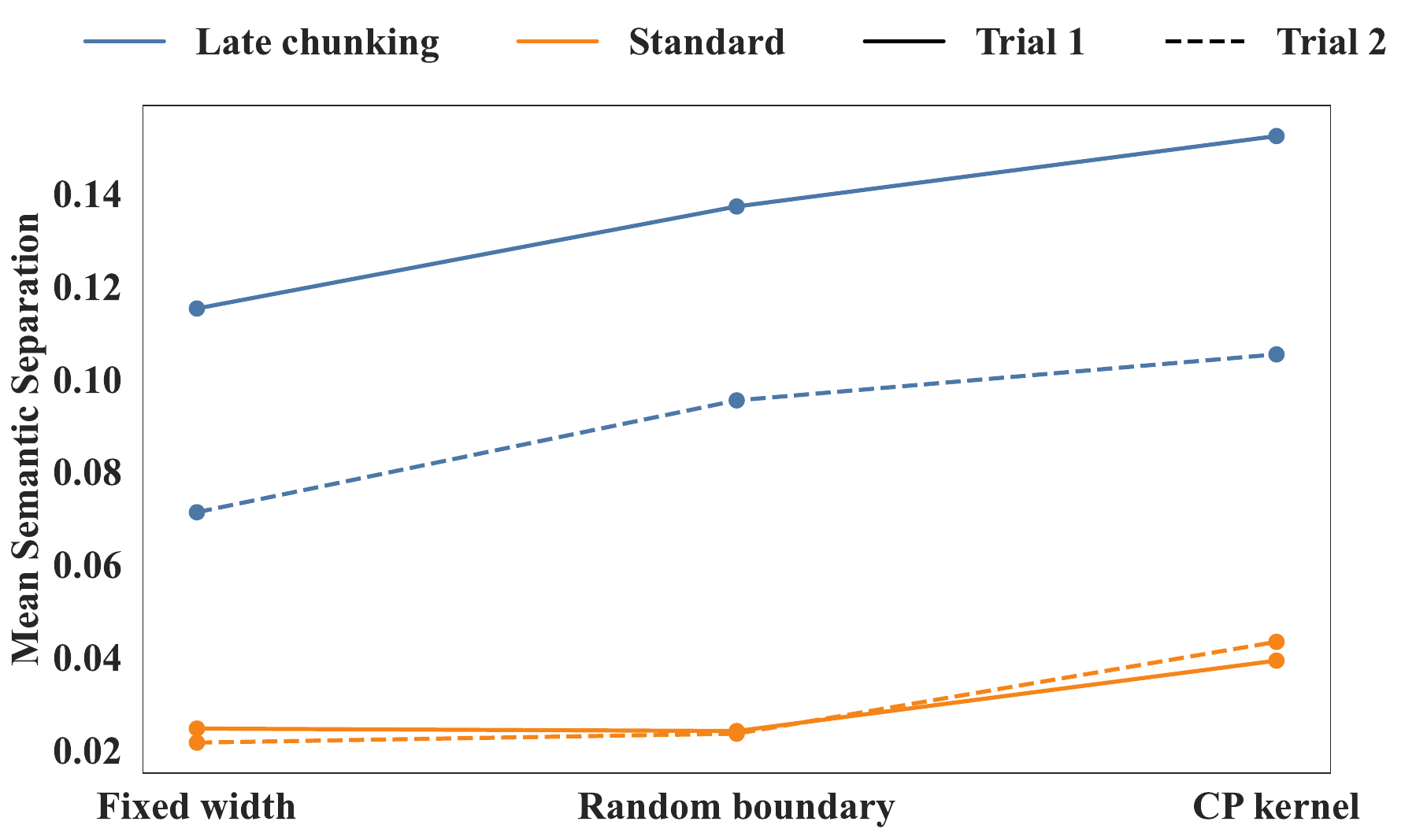}
    \caption{Semantic separation across segmentation methods and embedding modes.}
    \label{fig:baseline_semantic_combined}
\end{figure}

Under the final configuration using late-chunked \texttt{jina-v2-de}, mean pooling, a gamma multiplier of $\alpha=1$, and a penalty of $\beta=1$, Fig.~\ref{fig:baseline_semantic_combined} compares change-point kernel segmentation with fixed-width and random-boundary references using the same semantic-separation metric. For each embedding method, the phase count produced by change-point kernel segmentation defines the target number of phases for the reference methods, ensuring that all methods are compared under matched segmentation granularity. Change-point kernel segmentation achieves the highest semantic separation under late chunking, indicating greater internal semantic separation under the evaluation metric used here when contextual representation and adaptive boundary placement are combined. Fixed-width and random-boundary references preserve the same number of phases but do not use semantic changes to determine boundary locations.

\subsection{Phase Count, Timelines, and Interaction Profiles}
\label{subsec:phase_count_distribution_results}

\begin{figure}[!htbp]
    \centering
    \safeincludegraphics[width=0.5\columnwidth]{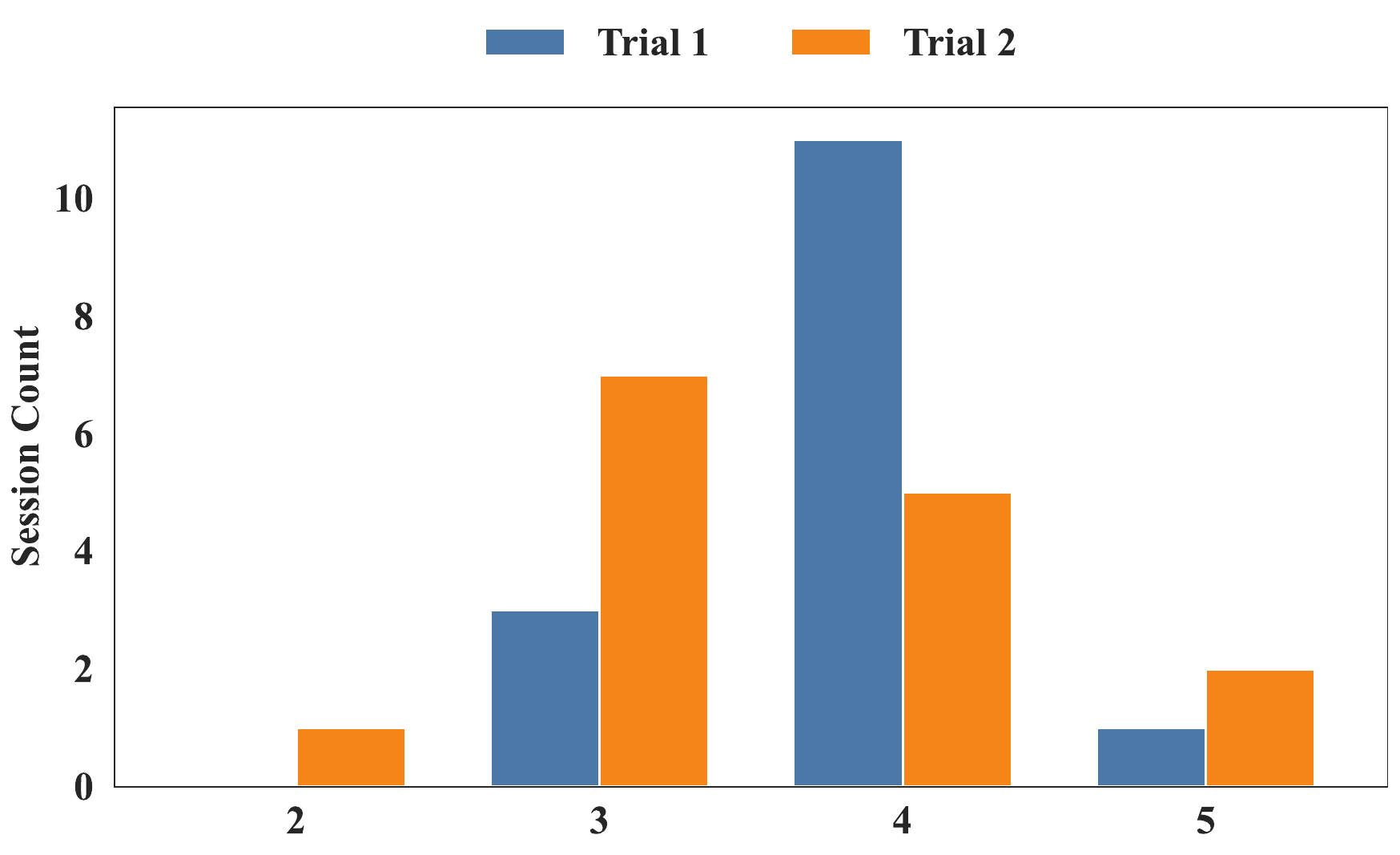}
    \caption{Distribution of detected phase counts across Trials~1 and~2.}
    \label{fig:phase_count_distribution_by_trial}
\end{figure}

The combined phase-structure and interaction results reveal descriptive differences in teamwork between the two trials. As shown in Fig.~\ref{fig:phase_count_distribution_by_trial}, Trial~1 is dominated by four-phase solutions, with such solutions observed in 11 of the 15 sessions, whereas three-phase solutions are most frequent in Trial~2, occurring in seven sessions. 

\begin{figure}[!htbp]
    \centering
    \safeincludegraphics[width=0.5\columnwidth]{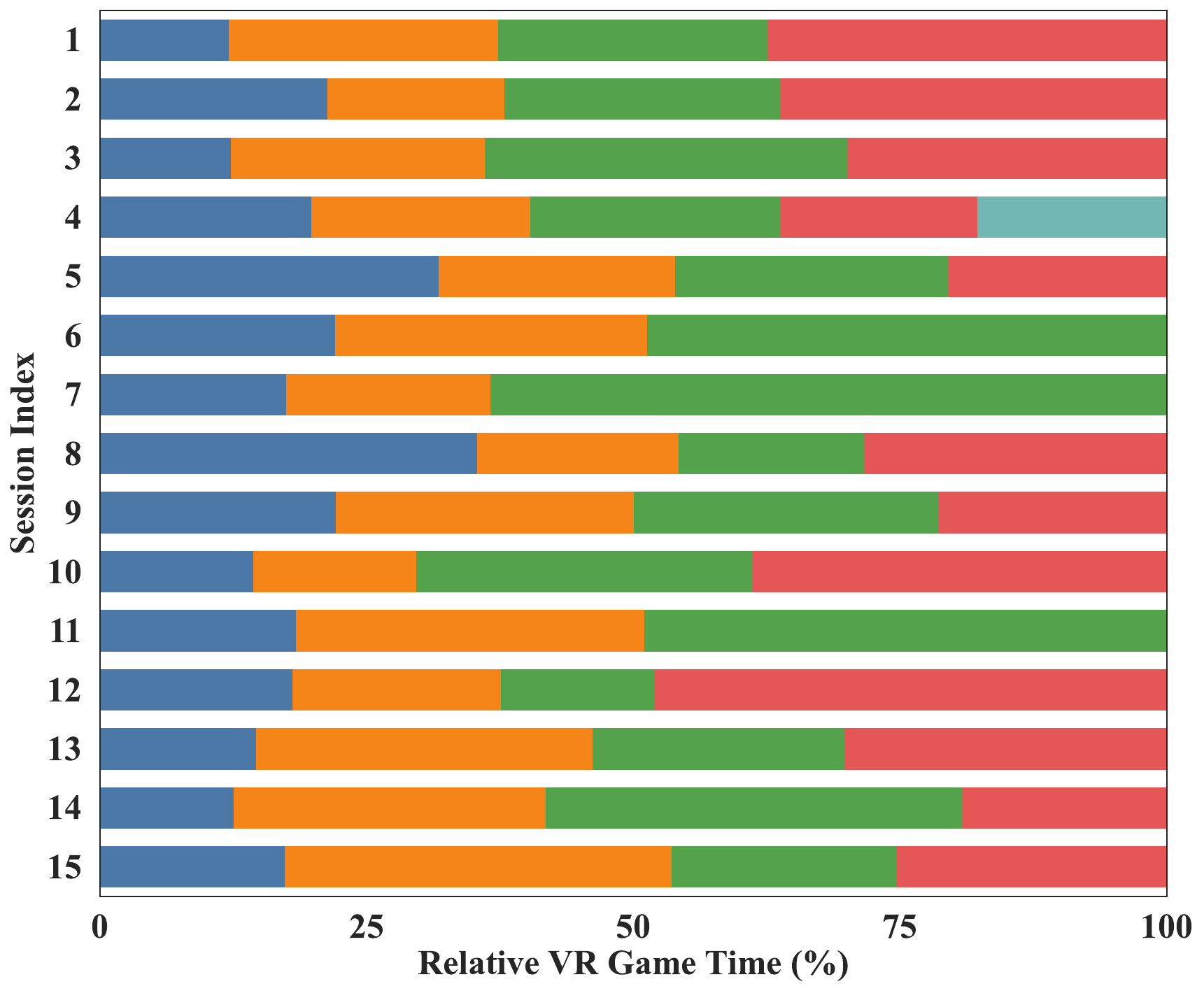}
    \caption{Normalized detected-phase timelines for Trial~1 sessions.}
    \label{fig:trial_1_phase_timeline}
\end{figure}

\begin{figure}[!htbp]
    \centering
    \safeincludegraphics[width=0.5\columnwidth]{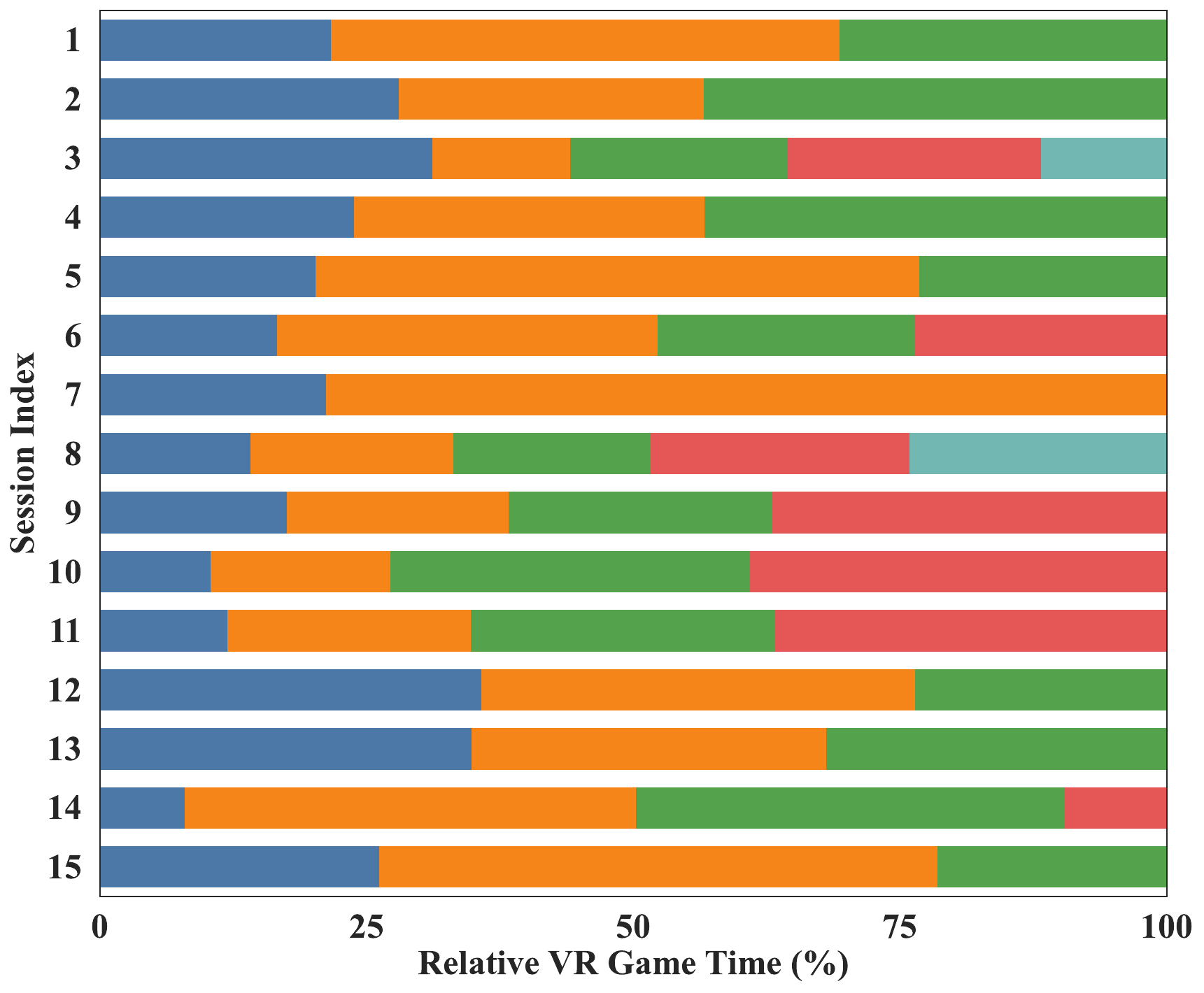}
    \caption{Normalized detected-phase timelines for Trial~2 sessions.}
    \label{fig:trial_2_phase_timeline}
\end{figure}

The timelines in Figs.~\ref{fig:trial_1_phase_timeline} and~\ref{fig:trial_2_phase_timeline} show the temporal order and relative duration of the detected phases for each team. Colors indicate ordinal phase positions from Phase~1 to Phase~5 within each session. The color-coded positions visualize the sequence of semantic changes and do not imply that phases with the same number have the same communicative or coordination meaning across teams. The specific phase meanings are examined in Section~\ref{sec:llm_phase_interpretation}. The team-specific boundary locations further show that the detected phases do not follow a fixed temporal template. Together, the phase-count and timeline results show that Trial~2 sessions more often contain fewer major changes in the semantic organization of team communication. Combined with the stronger task engagement observed at earlier ordinal phase positions in the interaction profiles, this pattern is consistent with more stable collaboration in Trial~2 after the initial task experience and interim reflection.

To examine whether the transcript-derived phase structure corresponds to task behavior, independently recorded interaction events are aligned with the detected phase intervals after segmentation. These events are not used to estimate the phase boundaries.

For phase \(p\) in session \(i\), the manipulation count is defined as
\begin{equation}
M_{ip}=G_{ip}+D_{ip},
\label{eq:manipulation_count}
\end{equation}
where \(G_{ip}\) and \(D_{ip}\) denote the numbers of object-grab and object-drop events assigned to that phase.

The manipulation rate is
\begin{equation}
R^{\mathrm{manip}}_{ip}
=
\frac{M_{ip}}{d_{ip}},
\label{eq:manipulation_rate}
\end{equation}
where \(d_{ip}\) is the phase duration in minutes. This measure represents task-action intensity.

The score rate is
\begin{equation}
R^{\mathrm{score}}_{ip}
=
\frac{Q_{ip}}{d_{ip}},
\label{eq:score_rate}
\end{equation}
where \(Q_{ip}\) is the number of scoring events in the phase. This measure represents scoring progress per minute.

Score per manipulation is
\begin{equation}
E_{ip}
=
\frac{Q_{ip}}{M_{ip}},
\qquad
M_{ip}>0.
\label{eq:score_per_manipulation}
\end{equation}
This measure represents how effectively object manipulations translate into successful scoring outcomes. When no manipulation event occurs within a phase, \(E_{ip}\) is treated as undefined rather than assigned a value of zero.

\begin{figure}[!htbp]
    \centering
    \safeincludegraphics[width=0.5\columnwidth]{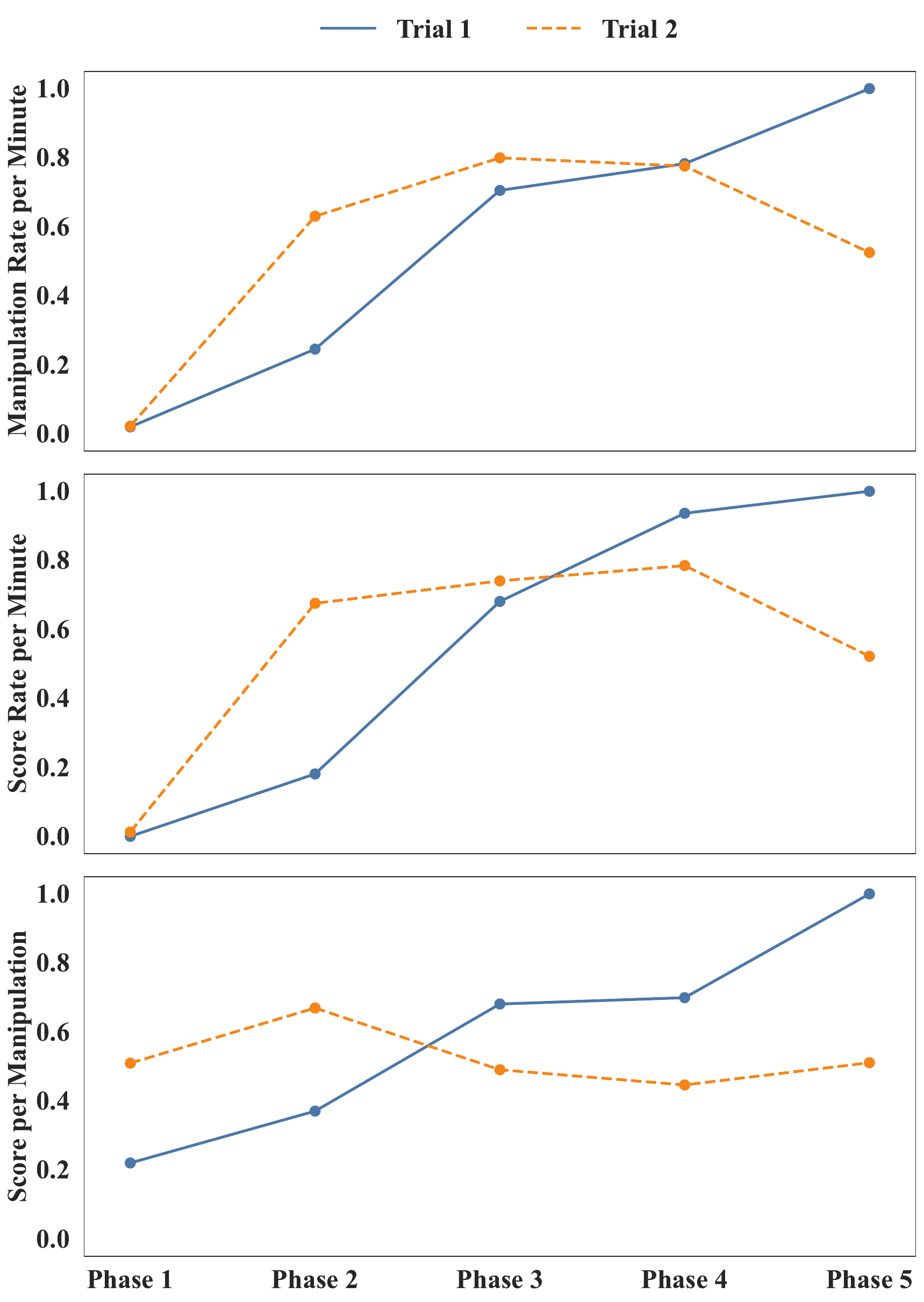}
    \caption{Phase-aligned interaction measures by ordinal phase position in Trials~1 and~2.}
    \label{fig:phase_level_interactive_action}
\end{figure}

The interaction profiles in Fig.~\ref{fig:phase_level_interactive_action} indicate stronger task engagement at earlier ordinal phase positions in Trial~2. Manipulation rate rises quickly and peaks in Phase~3, whereas Trial~1 shows a more gradual increase across the commonly observed phase positions. Score rate follows a similar pattern. In Trial~2, the score rate increases sharply by Phase~2 and remains comparatively high through Phases~3 and~4, whereas the score rate in Trial~1 increases more gradually. Score per manipulation also reaches a relatively high and stable level at earlier ordinal phase positions in Trial~2, suggesting that teams convert object manipulations into successful outcomes more consistently in the earlier detected phases. 

Taken together, the three measures suggest stronger task engagement, scoring progress, and more stable manipulation efficiency at earlier ordinal phase positions in Trial~2. This pattern is consistent with teams becoming familiar with the virtual environment, task procedure, and available objects during the first trial and applying that experience more effectively in the second.

The correspondence between the transcript-derived phase structure in Figs.~\ref{fig:phase_count_distribution_by_trial}--\ref{fig:trial_2_phase_timeline} and the independently recorded interaction profiles in Fig.~\ref{fig:phase_level_interactive_action} provides complementary behavioral evidence for the relevance of the detected phases. Because phase boundaries are estimated exclusively from transcript semantics and interaction measures are aligned only afterward, the observed correspondence indicates that the transcript-derived phases are associated with differences in collaborative task activity.

\section{Evidence-Grounded Phase Interpretation}
\label{sec:llm_phase_interpretation}

After CP-kernel segmentation, each fixed phase is interpreted using its time span, TF--IDF terms, NMF topic evidence, representative German transcript segments, and previous-phase context. A locally deployed LLM generates an initial English interpretation that is subsequently reviewed by humans. Each phase receives a reviewed phase-process label and interpreted NMF topic labels, while boundary placement remains unaffected by the interpretation step.

Table~\ref{tab:phase_position_pattern_summary} summarizes the most frequent phase-process labels by ordinal position and trial. Tables~\ref{tab:trial1_team_process_by_phase_csv} and~\ref{tab:trial2_team_process_by_phase_csv} show the phase sequences and supporting NMF topic evidence for all 15 teams in Trial~1 and Trial~2, allowing within-team changes across trials to be examined.

In Trial~1, Object/Resource Coordination and Player-Attribute Assignment are most frequent in the early phases, while Spatial Orientation and Localization becomes more prominent in Phases~3 and~4. At the aggregate level, this distribution is consistent with a shift from establishing player roles, identifying resources, and clarifying task requirements toward spatially grounded execution, and it corresponds to the gradual increase in manipulation and scoring shown in Fig.~\ref{fig:phase_level_interactive_action}.

In Trial~2, Player-Attribute Assignment is most frequent in Phase~1, followed by Mutual Visibility Check and Spatial Orientation and Localization in Phases~2 and~3. Together with the stronger task engagement observed at earlier ordinal phase positions in Fig.~\ref{fig:phase_level_interactive_action}, this pattern suggests that role establishment, shared visibility, and spatial relations are more prominent in the earlier detected phases of Trial~2 following the initial task experience and interim reflection.

Phase~5 is excluded from the cross-trial interpretation because it occurs in only one Trial~1 session and two Trial~2 sessions. The example sequences also show that teams do not follow a single fixed trajectory and that the same phase-process label can be supported by different NMF topic evidence.

\begin{table}[!htbp]
\centering
\caption{Most frequent reviewed phase-process labels at each ordinal phase position in Trials~1 and~2.}
\label{tab:phase_position_pattern_summary}
\scriptsize
\setlength{\tabcolsep}{2pt}
\renewcommand{\arraystretch}{1.08}
\begin{tabularx}{\columnwidth}{@{}P{0.1\columnwidth}P{0.44\columnwidth}P{0.44\columnwidth}@{}}
\toprule
Phase position & Trial~1 labels & Trial~2 labels \\
\midrule
Phase 1 & Object/Resource Coordination (5/15), Player-Attribute Assignment (4/15) & Player-Attribute Assignment (6/15), Player Identity Grounding (3/15) \\
Phase 2 & Player-Attribute Assignment (4/15), Object/Resource Coordination (3/15) & Mutual Visibility Check (5/15), Player-Attribute Assignment (5/15) \\
Phase 3 & Spatial Orientation and Localization (8/15), Player-Attribute Assignment (3/15) & Mutual Visibility Check (4/14), Spatial Orientation and Localization (4/14) \\
Phase 4 & Spatial Orientation and Localization (5/12), Player-Attribute Assignment (4/12) & Spatial Orientation and Localization (4/7) \\
Phase 5 & Player Identity Grounding (1/1) & Spatial Orientation and Localization (1/2), Object/Resource Coordination (1/2) \\
\bottomrule
\end{tabularx}
\end{table}

\begin{table*}[!htbp]
\centering
\caption{Reviewed team-process functions and supporting NMF topic labels across detected phases in Trial~1.}
\label{tab:trial1_team_process_by_phase_csv}
\tiny
\setlength{\tabcolsep}{0.5pt}
\renewcommand{\arraystretch}{1.04}
\begin{tabularx}{\textwidth}{@{}P{0.03\textwidth}P{0.194\textwidth}P{0.194\textwidth}P{0.194\textwidth}P{0.194\textwidth}P{0.194\textwidth}@{}}
\toprule
Index & Phase 1 & Phase 2 & Phase 3 & Phase 4 & Phase 5 \\
\midrule
ID01 & \parbox[t]{\linewidth}{Player-Attribute Assignment \\ {\tiny NMF: T1: Resource identification T2: Color identity verification}} & \parbox[t]{\linewidth}{Spatial Orientation and Localization \\ {\tiny NMF: T1: Numerical Sequence Coordination T2: Strategic and Empathic Approach T3: Spatial Re-orientation}} & \parbox[t]{\linewidth}{Mutual Visibility Check \\ {\tiny NMF: T1: Targeted visibility checks T2: Visual perception verification}} & \parbox[t]{\linewidth}{Spatial Orientation and Localization \\ {\tiny NMF: T1: Spatial Orientation via Gestures T2: Resource color identification}} & -- \\
ID02 & \parbox[t]{\linewidth}{Player-Attribute Assignment \\ {\tiny NMF: T1: Visual filter identification T2: Resource color identification T3: Visual Perception Calibration}} & \parbox[t]{\linewidth}{Player-Attribute Assignment \\ {\tiny NMF: T1: Movement Synchronization T2: Color identity declaration}} & \parbox[t]{\linewidth}{Player-Attribute Assignment \\ {\tiny NMF: T1: Avatar identification T2: Avatar visual identification}} & \parbox[t]{\linewidth}{Player-Attribute Assignment \\ {\tiny NMF: T1: Resource color identification T2: Resource Color Identification}} & -- \\
ID03 & \parbox[t]{\linewidth}{Object / Resource Coordination \\ {\tiny NMF: T1: Task failure acknowledgment T2: Tower construction planning}} & \parbox[t]{\linewidth}{Player-Attribute Assignment \\ {\tiny NMF: T1: Perceptual Color Verification T2: Error detection and confirmation T3: Spatial resource localization}} & \parbox[t]{\linewidth}{Spatial Orientation and Localization \\ {\tiny NMF: T1: Purple resource identification T2: Blue Resource Identification}} & \parbox[t]{\linewidth}{Spatial Orientation and Localization \\ {\tiny NMF: T1: Resource color identification T2: Resource visibility verification T3: Resource color identification}} & -- \\
ID04 & \parbox[t]{\linewidth}{Player-Attribute Assignment \\ {\tiny NMF: T1: Task procedure inquiry T2: Resource count verification}} & \parbox[t]{\linewidth}{Object / Resource Coordination \\ {\tiny NMF: T1: Task Execution Breakdown T2: Object manipulation attempts}} & \parbox[t]{\linewidth}{Spatial Orientation and Localization \\ {\tiny NMF: T1: Shared Visual Perception T2: Repeated Rejection of Incorrect Cues}} & \parbox[t]{\linewidth}{Player-Attribute Assignment \\ {\tiny NMF: T1: Green Resource Identification T2: Failed resource retrieval T3: Player identification queries}} & \parbox[t]{\linewidth}{Player Identity Grounding \\ {\tiny NMF: T1: Avatar color identification T2: Color identity verification T3: Directives for green guest location}} \\
ID05 & \parbox[t]{\linewidth}{Action Instruction and Turn Coordination \\ {\tiny NMF: T1: Action Synchronization T2: Fragmented Attention Signals T3: Expressions of Distress T4: Personal resource identification}} & \parbox[t]{\linewidth}{Player Identity Grounding \\ {\tiny NMF: T1: Execution Speed Regulation T2: Visual Grounding Struggle}} & \parbox[t]{\linewidth}{Spatial Orientation and Localization \\ {\tiny NMF: T1: Resource color identification T2: Turn-taking regulation}} & \parbox[t]{\linewidth}{Mutual Visibility Check \\ {\tiny NMF: T1: Visual Visibility Checks T2: Visual Grounding T3: Visual resource identification}} & -- \\
ID06 & \parbox[t]{\linewidth}{Player Identity Grounding \\ {\tiny NMF: T1: Identity verification T2: Expressions of Surprise T3: Collective Tower Construction}} & \parbox[t]{\linewidth}{Object / Resource Coordination \\ {\tiny NMF: T1: Resource Path Selection T2: Identity verification via color}} & \parbox[t]{\linewidth}{Player Identity Grounding \\ {\tiny NMF: T1: Color-based Identity Declaration T2: Mutual Visibility Establishment}} & -- & -- \\
ID07 & \parbox[t]{\linewidth}{Object / Resource Coordination \\ {\tiny NMF: T1: Spatial Orientation T2: Dice placement verification}} & \parbox[t]{\linewidth}{Repair and Re-grounding \\ {\tiny NMF: T1: Waiting for external cues T2: Exclusion Status Reporting}} & \parbox[t]{\linewidth}{Spatial Orientation and Localization \\ {\tiny NMF: T1: Resource location queries T2: Visual Grounding Confirmation T3: Visual resource identification}} & -- & -- \\
ID08 & \parbox[t]{\linewidth}{Object / Resource Coordination \\ {\tiny NMF: T1: Spatial orientation and identification T2: Resource Availability Negotiation}} & \parbox[t]{\linewidth}{Spatial Orientation and Localization \\ {\tiny NMF: T1: Color-location mapping T2: Resource location identification}} & \parbox[t]{\linewidth}{Player-Attribute Assignment \\ {\tiny NMF: T1: Visual property verification T2: Color-based resource identification}} & \parbox[t]{\linewidth}{Mutual Visibility Check \\ {\tiny NMF: T1: Resource color identification T2: Spatial location verification T3: Visual Contact Verification}} & -- \\
ID09 & \parbox[t]{\linewidth}{Mutual Visibility Check \\ {\tiny NMF: T1: Expressions of shock T2: Realization and acknowledgment T3: Visual Presence Verification}} & \parbox[t]{\linewidth}{Task-Constraint Grounding \\ {\tiny NMF: T1: Resource selection negotiation T2: Graspability Verification T3: Avatar Identity Verification T4: Repeated Rejection Signals}} & \parbox[t]{\linewidth}{Object / Resource Coordination \\ {\tiny NMF: T1: Spatial orientation of resources T2: Positive performance acknowledgment}} & \parbox[t]{\linewidth}{Object / Resource Coordination \\ {\tiny NMF: T1: Ownership identification T2: Member location verification T3: Resource Inventory Declaration T4: Graspability Verification}} & -- \\
ID10 & \parbox[t]{\linewidth}{Object / Resource Coordination \\ {\tiny NMF: T1: Object Location Grounding T2: Contact establishment}} & \parbox[t]{\linewidth}{Repair and Re-grounding \\ {\tiny NMF: T1: Exit and Visibility Loss T2: Expressions of Confusion}} & \parbox[t]{\linewidth}{Spatial Orientation and Localization \\ {\tiny NMF: T1: Purple resource location search T2: Color assignment verification}} & \parbox[t]{\linewidth}{Player-Attribute Assignment \\ {\tiny NMF: T1: Color resource identification T2: Identity verification queries T3: Spatial Proximity Coordination}} & -- \\
ID11 & \parbox[t]{\linewidth}{Player-Attribute Assignment \\ {\tiny NMF: T1: Permission assignment by color T2: Seeking opinions and clarification T3: Resource identification T4: Failed resource identification}} & \parbox[t]{\linewidth}{Player-Attribute Assignment \\ {\tiny NMF: T1: Resource color assignment T2: Resource identification queries}} & \parbox[t]{\linewidth}{Player-Attribute Assignment \\ {\tiny NMF: T1: Confirming purple resource assignment T2: Verbal color grounding}} & -- & -- \\
ID12 & \parbox[t]{\linewidth}{Player Identity Grounding \\ {\tiny NMF: T1: Resource identification T2: Spatial convergence coordination}} & \parbox[t]{\linewidth}{Player Identity Grounding \\ {\tiny NMF: T1: Avatar color identification T2: Color identity verification}} & \parbox[t]{\linewidth}{Spatial Orientation and Localization \\ {\tiny NMF: T1: Resource assignment queries T2: Movement coordination}} & \parbox[t]{\linewidth}{Spatial Orientation and Localization \\ {\tiny NMF: T1: Resource location queries T2: Resource attribute verification T3: Resource identification T4: Spatial Reorientation}} & -- \\
ID13 & \parbox[t]{\linewidth}{Player Identity Grounding \\ {\tiny NMF: T1: Blue resource identification T2: Mutual Orientation Struggle}} & \parbox[t]{\linewidth}{Player-Attribute Assignment \\ {\tiny NMF: T1: Resource Identity Confirmation T2: Identity declaration T3: Identity verification queries}} & \parbox[t]{\linewidth}{Spatial Orientation and Localization \\ {\tiny NMF: T1: Player Identity Identification T2: Red resource identification}} & \parbox[t]{\linewidth}{Player-Attribute Assignment \\ {\tiny NMF: T1: Avatar Identity Confirmation T2: Action Synchronization T3: Color-based role assignment}} & -- \\
ID14 & \parbox[t]{\linewidth}{Object / Resource Coordination \\ {\tiny NMF: T1: Spatial location verification T2: Visual resource selection T3: Resource identification repair T4: Visual Perception Verification}} & \parbox[t]{\linewidth}{Object / Resource Coordination \\ {\tiny NMF: T1: Resource identification by color T2: Resource Possession Confirmation}} & \parbox[t]{\linewidth}{Object / Resource Coordination \\ {\tiny NMF: T1: Action Synchronization T2: Explicit Waving Signaling T3: Team member identification}} & \parbox[t]{\linewidth}{Spatial Orientation and Localization \\ {\tiny NMF: T1: Grounding orange resource locations T2: Spatial Resource Identification}} & -- \\
ID15 & \parbox[t]{\linewidth}{Mutual Visibility Check \\ {\tiny NMF: T1: Visual perception confirmation T2: Spatial resource localization}} & \parbox[t]{\linewidth}{Mutual Visibility Check \\ {\tiny NMF: T1: Visual Grounding Verification T2: Initial spatial orientation T3: Yellow Resource Identification}} & \parbox[t]{\linewidth}{Spatial Orientation and Localization \\ {\tiny NMF: T1: Spatial location verification T2: Red Cube Identification}} & \parbox[t]{\linewidth}{Spatial Orientation and Localization \\ {\tiny NMF: T1: Resource identification T2: Commitment to immediate action T3: Hand gesture signaling T4: Resource color identification}} & -- \\
\bottomrule
\end{tabularx}
\end{table*}

\begin{table*}[!htbp]
\centering
\caption{Reviewed team-process functions and supporting NMF topic labels across detected phases in Trial~2.}
\label{tab:trial2_team_process_by_phase_csv}
\tiny
\setlength{\tabcolsep}{0.5pt}
\renewcommand{\arraystretch}{1.04}
\begin{tabularx}{\textwidth}{@{}P{0.03\textwidth}P{0.194\textwidth}P{0.194\textwidth}P{0.194\textwidth}P{0.194\textwidth}P{0.194\textwidth}@{}}
\toprule
Index & Phase 1 & Phase 2 & Phase 3 & Phase 4 & Phase 5 \\
\midrule
ID01 & \parbox[t]{\linewidth}{Player Identity Grounding \\ {\tiny NMF: T1: Avatar color identification T2: Identity declaration}} & \parbox[t]{\linewidth}{Mutual Visibility Check \\ {\tiny NMF: T1: Visual Perception Grounding T2: Resource identification via color}} & \parbox[t]{\linewidth}{Mutual Visibility Check \\ {\tiny NMF: T1: Visual confirmation of yellow resource T2: Resource location confirmation T3: Visual grounding statements}} & -- & -- \\
ID02 & \parbox[t]{\linewidth}{Spatial Orientation and Localization \\ {\tiny NMF: T1: Target location inquiry T2: Spatial resource identification}} & \parbox[t]{\linewidth}{Player-Attribute Assignment \\ {\tiny NMF: T1: Avatar Identity Verification T2: Avatar Identity Verification}} & \parbox[t]{\linewidth}{Player Identity Grounding \\ {\tiny NMF: T1: Color identity confirmation T2: Visual resource identification}} & -- & -- \\
ID03 & \parbox[t]{\linewidth}{Player-Attribute Assignment \\ {\tiny NMF: T1: Shared Visual Perception T2: Spatial reference confirmation}} & \parbox[t]{\linewidth}{Player-Attribute Assignment \\ {\tiny NMF: T1: Shared Visual Perception T2: Visual Color Identification}} & \parbox[t]{\linewidth}{Mutual Visibility Check \\ {\tiny NMF: T1: Visual Contact Establishment T2: Non-verbal visual signaling T3: Spatial orientation via color cues}} & \parbox[t]{\linewidth}{Spatial Orientation and Localization \\ {\tiny NMF: T1: Resource color identification T2: Avatar Movement Coordination T3: Interactional Acknowledgment T4: Spatial resource identification}} & \parbox[t]{\linewidth}{Spatial Orientation and Localization \\ {\tiny NMF: T1: Spatial resource orientation T2: Gestural signaling for attention T3: Red container identification}} \\
ID04 & \parbox[t]{\linewidth}{Player-Attribute Assignment \\ {\tiny NMF: T1: Resource identification and confirmation T2: Resource verification T3: Visual Grounding and Orientation T4: Visual perspective sharing}} & \parbox[t]{\linewidth}{Mutual Visibility Check \\ {\tiny NMF: T1: Resource Identification T2: Task execution feedback T3: Resource identification failure T4: Visual Occlusion Reporting}} & \parbox[t]{\linewidth}{Mutual Visibility Check \\ {\tiny NMF: T1: Resource identification T2: Green Resource Identification T3: Visual Grounding Verification}} & -- & -- \\
ID05 & \parbox[t]{\linewidth}{Spatial Orientation and Localization \\ {\tiny NMF: T1: pink gate identification T2: Yellow object identification}} & \parbox[t]{\linewidth}{Mutual Visibility Check \\ {\tiny NMF: T1: Visual Presence Verification T2: Requests for Pauses}} & \parbox[t]{\linewidth}{Mutual Visibility Check \\ {\tiny NMF: T1: Mutual Visual Grounding T2: Non-verbal gesture signaling}} & -- & -- \\
ID06 & \parbox[t]{\linewidth}{Mutual Visibility Check \\ {\tiny NMF: T1: Resource location identification T2: Spatial Orientation and Identification T3: Visual Contact Establishment T4: Visual Grounding Confirmation}} & \parbox[t]{\linewidth}{Mutual Visibility Check \\ {\tiny NMF: T1: Visual Grounding Confirmation T2: Task execution feedback}} & \parbox[t]{\linewidth}{Object / Resource Coordination \\ {\tiny NMF: T1: Resource need inquiry T2: Resource identifier verification}} & \parbox[t]{\linewidth}{Object / Resource Coordination \\ {\tiny NMF: T1: Resource identification T2: Identity Grounding via Name}} & -- \\
ID07 & \parbox[t]{\linewidth}{Mutual Visibility Check \\ {\tiny NMF: T1: Mutual Visibility Verification T2: Spatial resource localization T3: Resource acquisition confirmation}} & \parbox[t]{\linewidth}{Spatial Orientation and Localization \\ {\tiny NMF: T1: Visual Grounding T2: Spatial boundary definition}} & -- & -- & -- \\
ID08 & \parbox[t]{\linewidth}{Player-Attribute Assignment \\ {\tiny NMF: T1: pink resource identification T2: Color identity declaration T3: Spatial orientation check T4: Blue Zügel Resource Identification}} & \parbox[t]{\linewidth}{Player-Attribute Assignment \\ {\tiny NMF: T1: orange Resource Identification T2: Resource assignment confirmation}} & \parbox[t]{\linewidth}{Player Identity Grounding \\ {\tiny NMF: T1: Green resource identification T2: Visual Grounding Confirmation T3: Resource attribute identification T4: Resource assignment verification}} & \parbox[t]{\linewidth}{Spatial Orientation and Localization \\ {\tiny NMF: T1: Resource location assignment T2: Spatial alignment invitation}} & \parbox[t]{\linewidth}{Object / Resource Coordination \\ {\tiny NMF: T1: Resource color assignment T2: Resource acquisition confirmation T3: Movement Coordination}} \\
ID09 & \parbox[t]{\linewidth}{Mutual Visibility Check \\ {\tiny NMF: T1: Visual Perception Verification T2: Blue cube identification}} & \parbox[t]{\linewidth}{Spatial Orientation and Localization \\ {\tiny NMF: T1: Resource location queries T2: Visual observation reporting}} & \parbox[t]{\linewidth}{Spatial Orientation and Localization \\ {\tiny NMF: T1: Resource color identification T2: Visual resource orientation}} & \parbox[t]{\linewidth}{Mutual Visibility Check \\ {\tiny NMF: T1: Color identity confirmation T2: Visual Presence Verification}} & -- \\
ID10 & \parbox[t]{\linewidth}{Player-Attribute Assignment \\ {\tiny NMF: T1: Spatial resource orientation T2: Color-based Identity Verification}} & \parbox[t]{\linewidth}{Spatial Orientation and Localization \\ {\tiny NMF: T1: Resource identification T2: Visual Reference Grounding}} & \parbox[t]{\linewidth}{Player Identity Grounding \\ {\tiny NMF: T1: Identity and Color Mapping T2: Color identity confirmation T3: Green resource identification T4: Avatar identification}} & \parbox[t]{\linewidth}{Player-Attribute Assignment \\ {\tiny NMF: T1: Resource assignment by color T2: Color-based resource identification}} & -- \\
ID11 & \parbox[t]{\linewidth}{Player-Attribute Assignment \\ {\tiny NMF: T1: Visual object identification T2: Resource color identification}} & \parbox[t]{\linewidth}{Player-Attribute Assignment \\ {\tiny NMF: T1: Player Identity Verification T2: Role identification via color}} & \parbox[t]{\linewidth}{Spatial Orientation and Localization \\ {\tiny NMF: T1: Resource color identification T2: Resource holder identification}} & \parbox[t]{\linewidth}{Spatial Orientation and Localization \\ {\tiny NMF: T1: Resource color identification T2: Blue container identification T3: Resource location identification}} & -- \\
ID12 & \parbox[t]{\linewidth}{Player Identity Grounding \\ {\tiny NMF: T1: Avatar identification T2: Green resource identification}} & \parbox[t]{\linewidth}{Spatial Orientation and Localization \\ {\tiny NMF: T1: Resource location queries T2: Identity confirmation T3: Resource color identification}} & \parbox[t]{\linewidth}{Spatial Orientation and Localization \\ {\tiny NMF: T1: Resource location queries T2: Resource location identification}} & -- & -- \\
ID13 & \parbox[t]{\linewidth}{Player-Attribute Assignment \\ {\tiny NMF: T1: Avatar identification by color T2: Color identity identification}} & \parbox[t]{\linewidth}{Player-Attribute Assignment \\ {\tiny NMF: T1: Resource identification via color T2: Temporal Coordination T3: Grounding Purple Player T4: Personal resource identification}} & \parbox[t]{\linewidth}{Player-Attribute Assignment \\ {\tiny NMF: T1: Blue task completion confirmation T2: Movement synchronization}} & -- & -- \\
ID14 & \parbox[t]{\linewidth}{Player Identity Grounding \\ {\tiny NMF: T1: Resource identification T2: Visual resource identification}} & \parbox[t]{\linewidth}{Mutual Visibility Check \\ {\tiny NMF: T1: Visual Grounding T2: Team Identity Identification T3: Spatial Identity Verification T4: Visual resource identification}} & \parbox[t]{\linewidth}{Object / Resource Coordination \\ {\tiny NMF: T1: pink Resource Identification T2: Member location queries}} & \parbox[t]{\linewidth}{Spatial Orientation and Localization \\ {\tiny NMF: T1: Grounding orange gate location T2: Location confirmation}} & -- \\
ID15 & \parbox[t]{\linewidth}{Spatial Orientation and Localization \\ {\tiny NMF: T1: Visual Object Verification T2: Resource identification queries T3: Visual resource localization}} & \parbox[t]{\linewidth}{Spatial Orientation and Localization \\ {\tiny NMF: T1: Spatial orientation T2: Spatial resource orientation}} & \parbox[t]{\linewidth}{Spatial Orientation and Localization \\ {\tiny NMF: T1: Resource acquisition confirmation T2: Resource color identification}} & -- & -- \\
\bottomrule
\end{tabularx}
\end{table*}

\section{Discussion, Limitations, Conclusion, and Ethics}
\label{sec:discussion_limitations_conclusion_ethics}

This study presents a phase-centered framework for detecting and interpreting dynamic team-process phases from timestamped collaborative VR dialogue. It combines context-aware transcript representations, adaptive semantic segmentation, evidence-grounded LLM-assisted interpretation, and phase-aligned interaction analysis. Because boundaries are fixed before interpretation, the assigned labels cannot influence their placement. Among the tested configurations, the configuration combining late-chunked \texttt{jina-v2-de}, mean pooling, and CP-kernel segmentation achieves the highest semantic separation. The main contribution is the detection of variable-length semantic phases without relying on fixed temporal windows or trial-level summaries. Each phase remains traceable to the original transcript segments, TF--IDF terms, and NMF topic evidence, allowing the resulting labels to be reviewed against their underlying evidence. The LLM supports this process by generating initial candidate labels, while final interpretations remain subject to human review.

The detected trajectories show how team communication shifts across grounding, orientation, coordination, and task execution. Independently recorded interaction and performance logs are aligned only after segmentation, allowing the transcript-derived phases to be examined in relation to corresponding task activity. This may support future after-action review and phase-specific feedback in collaborative training.

Several limitations remain. Semantic separation is an internal evaluation metric because boundary detection and evaluation use the same embedding representation. In contrast, the independently recorded interaction and performance logs provide complementary behavioral evidence that is aligned only after segmentation. The reviewed labels should nevertheless be understood as evidence-grounded interpretations rather than direct measurements of latent psychological states or replacements for independent behavioral coding. Future work should compare the detected phases with trainer ratings, participant self-reports, independent behavioral annotations, physiological signals, and broader performance outcomes.

The workflow also raises data-protection requirements because audio, transcripts, timestamps, and speaker information may contain identifiable data. Relevant safeguards include anonymization, data minimization, secure storage, local processing where possible, and continued human oversight of LLM-assisted interpretation. The framework is intended for research and training analysis rather than for ranking or diagnosing individual participants.

In conclusion, the framework transforms timestamped VR dialogue into interpretable semantic phase trajectories that remain traceable to transcript evidence and can be examined alongside task-action profiles. It provides a reproducible basis for analyzing how team communication and coordination change over time in collaborative human--machine systems.



\section*{Conflict of Interest}

The authors declare that they have no conflicts of interest.

\bibliographystyle{IEEEtran}
\bibliography{references}

\end{document}